\documentclass{article}

\usepackage[preprint,nonatbib]{neurips_2018}

\usepackage[utf8]{inputenc} 
\usepackage[T1]{fontenc}    
\usepackage{hyperref}       
\usepackage{url}            
\usepackage{booktabs}       
\usepackage{amsfonts}       
\usepackage{nicefrac}       
\usepackage{microtype}      

\usepackage{latexsym}
\usepackage[fleqn]{amsmath}
\usepackage{amssymb}
\usepackage[ruled,vlined,linesnumbered]{algorithm2e}
\usepackage{graphicx}

\title{On-line non-overlapping camera calibration net}

\author{%
Zhao Fangda,
Toru Tamaki,
Takio Kurita,
Bisser Raytchev,
Kazufumi Kaneda\\
Hiroshima University\\
Hiroshima 739-8527 Japan
}

\begin{document}

\maketitle

\begin{abstract}
We propose an easy-to-use non-overlapping camera calibration method.
First, successive images are fed to a PoseNet-based network to obtain ego-motion  of cameras between frames.
Next, the pose between cameras are estimated.
Instead of using a batch method, we propose an on-line method of the inter-camera pose estimation.
Furthermore, we implement the entire procedure on a computation graph.
Experiments with simulations and the KITTI dataset show the proposed method to be effective in simulation.
\end{abstract}

\section{Introduction}

Camera calibration is one of fundamental task in computer vision and has been studied over decades, while it is still a popular topic. Recently, calibration methods have been proposed for cameras that do not share the field of views (FOVs), which are called non-overlapping calibration. In this case, standard stereo camera calibration methods do not work.
For this task, many methods have been proposed
such as SLAM-based \cite{Ataer-cansizoglu2014,Carrera2011}, 
mirror-based \cite{Agrawal2013,Kumar2008,Takahashi2012,takahashi2016}, 
tracking-based \cite{Lamprecht2007,Anjum2007}, 
trajectory-based \cite{lebraly132010calibration,esquivel2007calibration},
and AR marker-based \cite{Zhao2016,Zhao2018}.

However, these approaches are off-line or computationally demanding.
The use of mirrors \cite{Agrawal2013,Kumar2008,Takahashi2012,takahashi2016} and AR makers \cite{Zhao2016,Zhao2018} enforces the calibration to be done before moving the calibration rig or driving cars (in case of car-mounted cameras). 
Trajectory-based methods \cite{lebraly132010calibration,esquivel2007calibration} takes camera trajectory data as a batch to process after shooting videos with the cameras.
SLAM-based methods \cite{Ataer-cansizoglu2014,Carrera2011} may work on-line, however a computation resources is usually required.

In this paper, we propose an on-line calibration method for non-overlapping cameras. Assuming the inter-camera pose being fixed, the proposed method continuously takes video frames as input and computes the inter-camera pose at each frame. This makes the calibration easy-to-use in practical situations. Our contributions are:
\begin{itemize}
\item First, we develop an on-line method for estimating inter-camera pose by extending an existing batch trajectory-based method.
\item We combine a PoseNet-based ego-motion network with our on-line pose update scheme.
\item Finally, we implement the entire procedure on a computation graph, that is, 
the network consists of the ego-motion part and inter-camera pose update part.
\end{itemize}

\section{Proposed method}

\subsection{Inter-camera pose estimation}

Here we briefly review a trajectory-based method \cite{Esquivel2007}
on which our proposed method is based.
Without loss of generality, we assume that there are two cameras
fixed on the same rig (hence do not move),
one is called master and the other is slave, hereafter we call them
camera 0 and 1, respectively.
Fixing the coordinate system of camera 0,
the problem is to estimate the coordinate system of camera 1,
represented by rotation matrix $\Delta R$ and translation $\Delta T$. In addition, the scale difference $\Delta \lambda$ between two cameras is also estimated; that is, two camera coordinates
are related by a similar transformation.

To this end, two sequences of camera poses are obtained
by moving the camera rig with the two cameras,
and estimating camera motions with respect to their initial positions.
More specifically,
$R_t^i$ and $T_t^i$ denote a pose of camera $i$ at time $t$,
and $R_0^0 = I$ and $T_0^0 = 0$,
and $R_0^1 = \Delta R$ and $T_0^1 = \Delta T$.
Furthere we denote 
$q_t^i$ as the quaternion of the rotation $R_t^i$,
and 
$A_t^i$ as the matrix composed of the quaternions
at initial frame $t=0$ and the current frame $t$
defined by 
\begin{equation*}
A_t^i =
\begin{pmatrix}
w_t^0 - w_t^i & -x_t^0 + x_t^i & -y_t^0 + y_t^i & -z_t^0 + z_t^i \\
x_t^0 - x_t^i &  w_t^0 - w_t^i & -z_t^0 - z_t^i &  y_t^0 + y_t^i \\
y_t^0 - y_t^i &  z_t^0 + z_t^i &  w_t^0 - w_t^i & -x_t^0 - x_t^i \\
z_t^0 - z_t^i & -y_t^0 - y_t^i &  x_t^0 + x_t^i &  w_t^0 - w_t^i \\
\end{pmatrix}
\end{equation*}
for $q_t^i = (w_t^i, x_t^i, y_t^i, z_t^i)$.

Given camera poses $\{R_t^i, T_t^i\} \ (t=1,2, \ldots, N)$ over $N$ frames,
rotation between cameras $\Delta q$ (quaternion of $\Delta R$)
is obtained by solving
\begin{equation}
\min_{\Delta q} \sum_t \| A_t^1 \Delta q \|^2.
\label{eq:R_estimation}
\end{equation}

Once rotation $\Delta R$ is recovered,
then translation $\Delta T$ and scale difference $\Delta \lambda$
is obtained by solving
\begin{equation}
\min_{\Delta x}
\sum_t \| B_t^1 \Delta x - T_t^0 \|^2,
\quad
\Delta x =
\begin{pmatrix}
\Delta T \\
\Delta \lambda
\end{pmatrix}
\label{eq:T_estimation}
\end{equation}
where
$B_t^i = (I - R_t^0, \Delta R T_t^i)$.

\subsection{PoseNet and Ego-motion estimation}

The trajectory-based method \cite{Esquivel2007} needs camera poses at each time step.
Here we use a PoseNet-based ego-motion estimation method.
PoseNet \cite{Kendall2015} is a network to estimate 6D camera pose by matching
the current video frame with learned 3D scenes. In this sense, this is a camera re-localization
because it can not deal with new scenes that are not seen during training the network.
It has been extended to an ego-motion estimation network \cite{Zhou2017}
which takes successive video frames to compute
the depth of the scene as well as the camera motion between the frames (i.e., ego-motion).

A straightforward way might be the use of the ego-motion results to perform 
the trajectory-based method. However, this naive idea has the following drawbacks.
First, this is a batch way because a specified number of (i.e., 200) frames are necessary to
compute the inter-camera pose. This needs to wait for hundreds of frames to obtain a result,
and not good when we want to adjust, or temporally attach, the cameras on the rig.
Second, this batch way is not useful when we want to compute the entire procedure on a GPU.
Instead, our method is on-line, good for the adjustable or temporal cameras,
and updates parameters on the same computation graph with the PoseNet-based ego-motion estimation.

\begin{figure}[t]
\centering
\includegraphics[width=\linewidth]{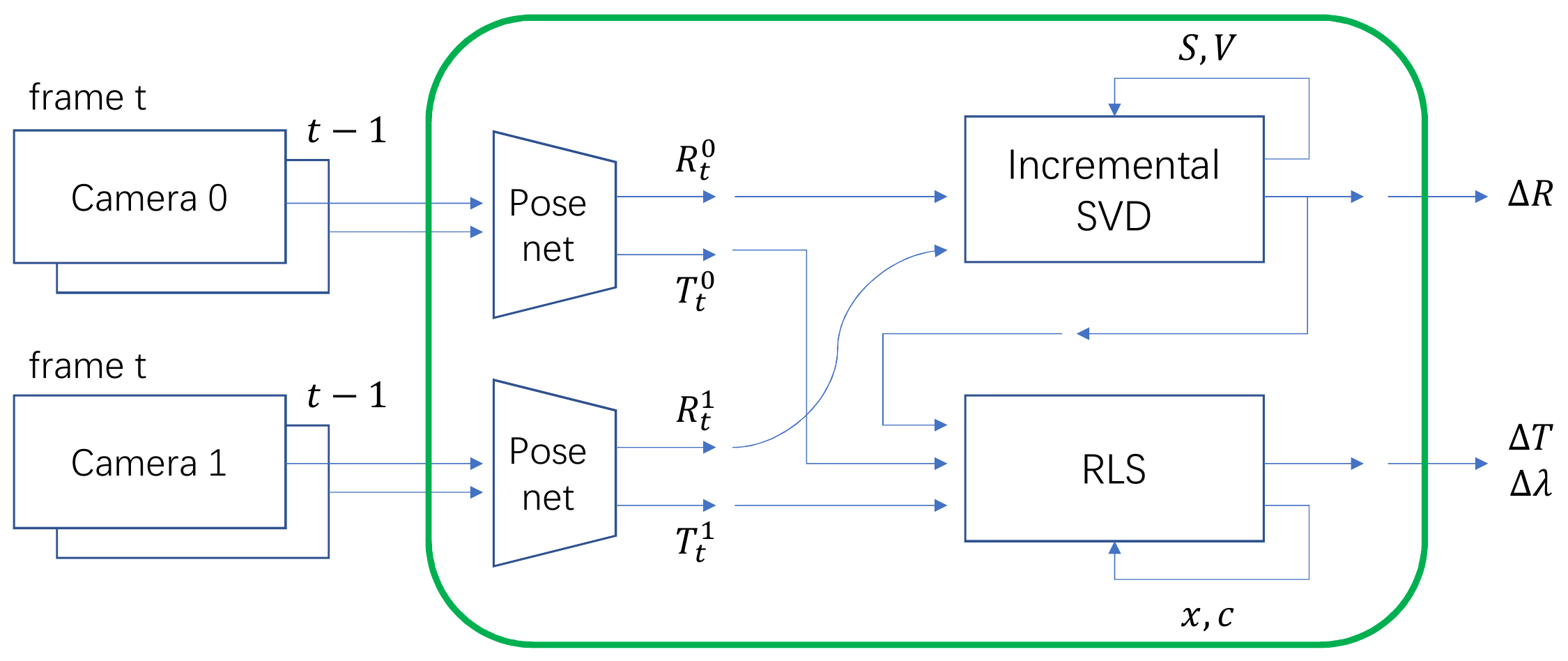}
\caption{Overview of the proposed method}
\label{fig:overview}
\end{figure}

\subsection{Incremental estimation}

Batch problems (\ref{eq:R_estimation})
and (\ref{eq:T_estimation})
can solved with SVD and least-square solver, respectively.
Here we propose to incrementally update the inter-pose
at each time.

\subsubsection{incremental SVD}

To solve problem (\ref{eq:R_estimation}),
we propose to use incremental SVD \cite{Bunch1978,Gu1994,Zha1999}.
First, SVD of matrix $A$ is given as $U S V^T$.
Then, new matrix $B$ is added by stacking them vertically
as $(A^T B^T)^T$.
Incremental SVD computes SVD of 
$(A^T B^T)^T$ by using $U S V^T = A$ and $B$ as follows.
\begin{enumerate}
\item Perform QL factorization $QL = (I - V V^T) B^T$

\item Perform SVD
$
\tilde{U}, \tilde{S}, \tilde{V}^T =
\begin{pmatrix}
S & 0 \\
B V & L
\end{pmatrix}
$

\item SVD of $(A^T B^T)^T$ is given as
$
\begin{pmatrix}
U & 0 \\
0 & I
\end{pmatrix}
\tilde{S}
\begin{pmatrix}
[V Q] \tilde{V}
\end{pmatrix}
$
\end{enumerate}

In our problem, $A$ is square and therefore $V$ is orthogonal,
which makes the update rules much simpler.

Furthermore, we can rewrite problem (\ref{eq:R_estimation})
as $(A_1^{1T}, A_2^{1T}, \ldots, A_N^{1T})^T \Delta q = 0$,
and the solution is the right singular vector $v_4$
corresponding the least singular value.
Therefore, we can discard the left singular vectors $U$
and keep the right singular vectors $V$ only.

Algorithm \ref{Alg:incremental} shows
our method for computing the right singular vectors of matrix
$(A_1^{1T}, A_2^{1T}, \ldots, A_N^{1T})^T$ recursively.

\subsubsection{RLS}

To solve problem (\ref{eq:T_estimation}),
we use recursive least squares (RLS) \cite{Haykin1996,Hayes1996,Sayed2003}.
Particularly, given $\{(B_t^1, T_t^0)\}_{t=1}^N$, 
we derive the following exponentiated block RSL 
with forgetting factor $\lambda$ as follows;
\begin{align}
\Gamma_t &= (I + \lambda^{-1} {B_t^1}^T C_{t-1} B_t^1)^{-1} \\
G_t &= \lambda^{-1} C_{t-1} {B_t^1}^T \Gamma_t \\
\Delta x_{t} &= \Delta x_{t-1} + G_t (T_t^0 - B_t^1 \Delta x_{t-1}) \\
C_t &= \lambda^{-1} C_{t-1} + G_t \Gamma_t^{-1} G_t^T,
\end{align}
with the initialization of 
$\Delta x_0$ to be a zero vector,
and $C_0$ to be an identity matrix.

\subsection{Proposed on-line algorithm}

The proposed method shown in Algorithm \ref{Alg:online}
combines the incremental SVD and RLS.
It takes video frames of both cameras
and estimate the inter-camera pose 
at each time.

Both the incremental SVD and RLS are recursive
but not iterative, which means that solutions are exact.
Our method is however not exactly the same with the batch method \cite{Esquivel2007}
because
$B_t^i$ in RLS includes a temporal solution of rotation $\Delta R$.

\begin{algorithm}[t]

\KwIn{$A_1, A_2, \ldots, A_N$}
\KwOut{Solution quaternion $q$}

$U S V^T = A_1$ \tcp*{initial SVD}

\For{$t = 2$ \KwTo $N$}{

$
\tilde{U} \tilde{S} \tilde{V}^T =
\begin{pmatrix}
S \\
A_t V
\end{pmatrix}
$ \tcp*{SVD}
 
$S = \tilde{S}$ \tcp*{update S}

$V = V \tilde{V}$ \tcp*{update V}
}

return $v_4$ as $q$ \tcp*{vector of least singular value}

\caption{Incremental SVD algorithm.}
\label{Alg:incremental}
\end{algorithm}

\begin{algorithm}[t]
\SetKw{KwInit}{Init: }
\KwInit{$S, V, C, x, \lambda$}

\KwIn{frames $I_t^0, I_{t-1}^0, I_t^1, I_{t-1}^1$}
\KwOut{$\Delta R, \Delta T, \Delta \lambda$}

\tcp{Ego-motion}

$R_t^0, T_t^0 = \mathrm{PoseNet}(I_t^0, I_{t-1}^0)$ \tcp*{first camera 0}

$R_t^1, T_t^1 = \mathrm{PoseNet}(I_t^1, I_{t-1}^1)$ \tcp*{second camera 1}

\tcp{Rotation with incremental SVD}

$A \leftarrow R_t^0, R_t^1$

$
\tilde{U} \tilde{S} \tilde{V}^T =
\begin{pmatrix}
S \\
A V
\end{pmatrix}
$ \tcp*{SVD}
 
$S = \tilde{S}; \ V = V \tilde{V}$ \tcp*{update S and V}

$\Delta R \leftarrow V$ \tcp*{Rotation between cameras}

\tcp{translation; RLS with forgetting factor}

$B \leftarrow R_t^0, \Delta R, T_t^1; \ b = T_t^0$

$\Gamma = (I + \lambda^{-1} B^T C B)^{-1}; \
G = \lambda^{-1} C B^T \Gamma$

$x = x + G (b - B x)$ \tcp*{update x}

$C = \lambda^{-1} C + G \Gamma^{-1} G^T$

$\Delta T, \Delta\lambda \leftarrow x$ \tcp*{Translation and scale}

return $\Delta R, \Delta T, \Delta\lambda$

\caption{On-line algorithm. Arrows indicate some trivial transformations.}
\label{Alg:online}
\end{algorithm}

As we use a PoseNet-based ego-motion estimation net,
it is reasonable to implement the on-line estimation
on the same computation graph on a GPU.
This reduces the memory transfer between GPUs and CPUs.
The overview of the proposed method is shown 
in Figure \ref{fig:overview}.
Note that in our current prototype implementation
two parts are separated due to some problems.

\section{Experimental results}

\subsection{Numerical simulation}

\begin{figure}[t]
\centering
\includegraphics[width=.6\linewidth]{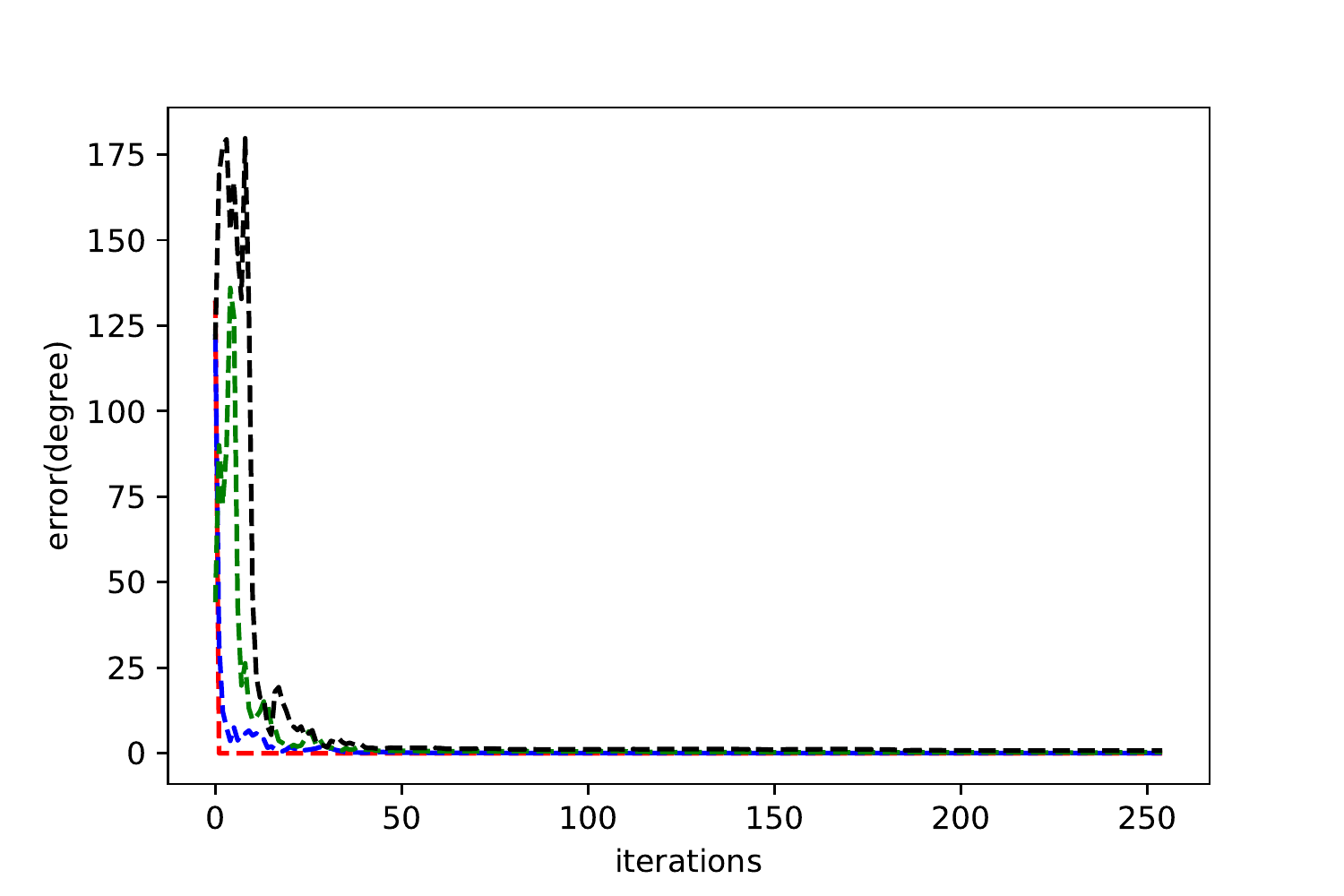}\\
\includegraphics[width=.6\linewidth]{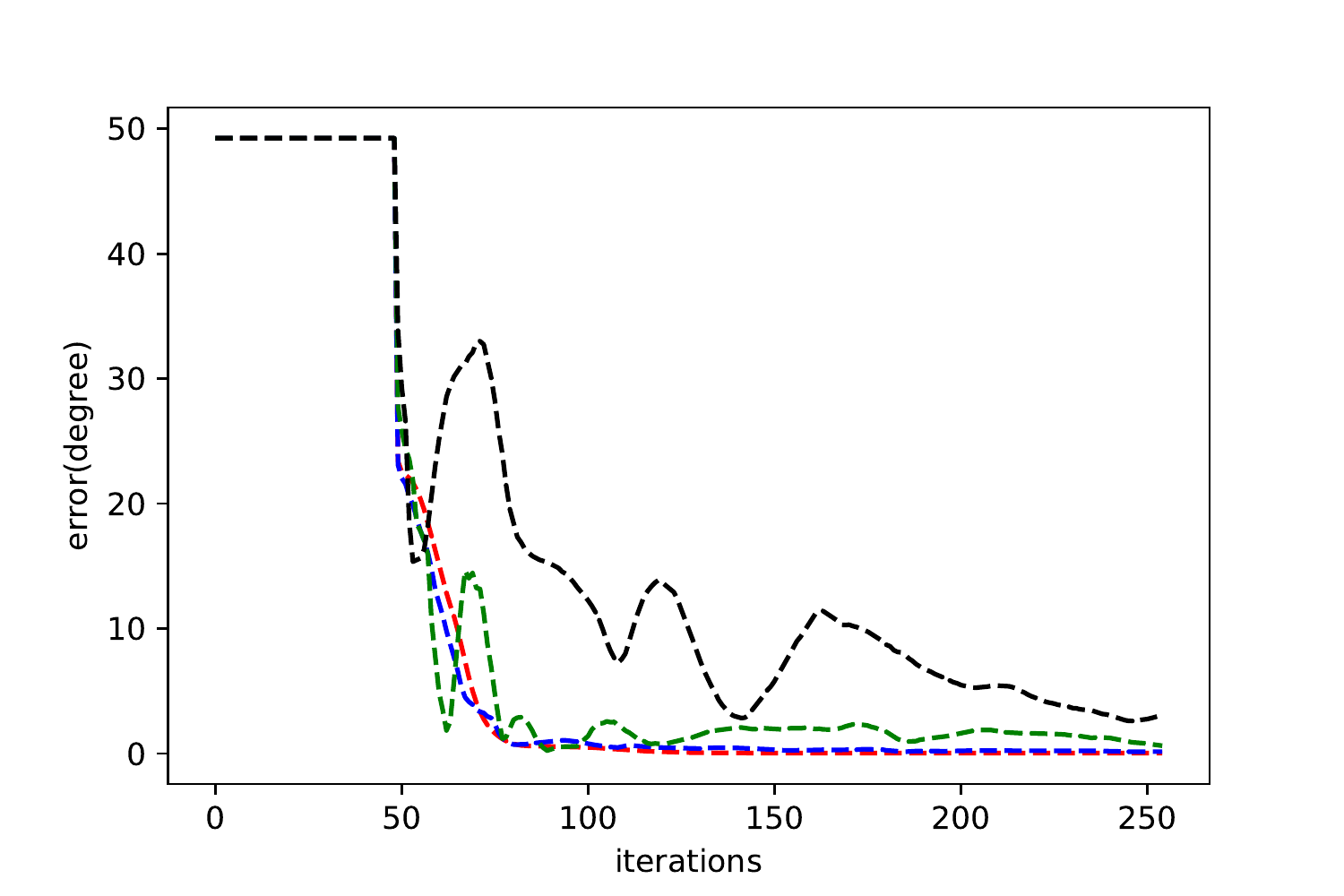}

\caption{Error of angles (in deg) in estimating (top) rotation axes and (bottom) translation vectors over iterations. Different curves show results with different amount of noise added to poses.}
\label{fig:error_simulation}
\end{figure}

First, we evaluated the proposed method with synthetic
trajectories. Two camera poses were randomly generated (but with a unit baseline length due to the scale ambiguity) and fixed on 
a rig, then a trajectory of the rig over 128 frames was generated by using a linear dynamic system.
We added Gaussian noise with zero mean and different stds (noise levels)
to rotation and translation of each frame independently.
Figure \ref{fig:error_simulation} shows relative errors of estimated inter-camera rotation and translation.

We see that errors degrease as iteration increases, and final errors (at the last iteration) becomes larger when the noise level is large. In rotation estimation, errors in first several frames are large but decreases rapidly. However this effect remains in translation estimation even after a large number of iterations. Therefore we postponed and started translation estimation at 60 iterations. This effectively reduces the error in translation, and results are acceptable in practical situations; translation errors about 0.25 with respect to the baseline length of 1.

\subsection{Real video frames}

\begin{figure}[t]
\centering
\includegraphics[width=.6\linewidth]{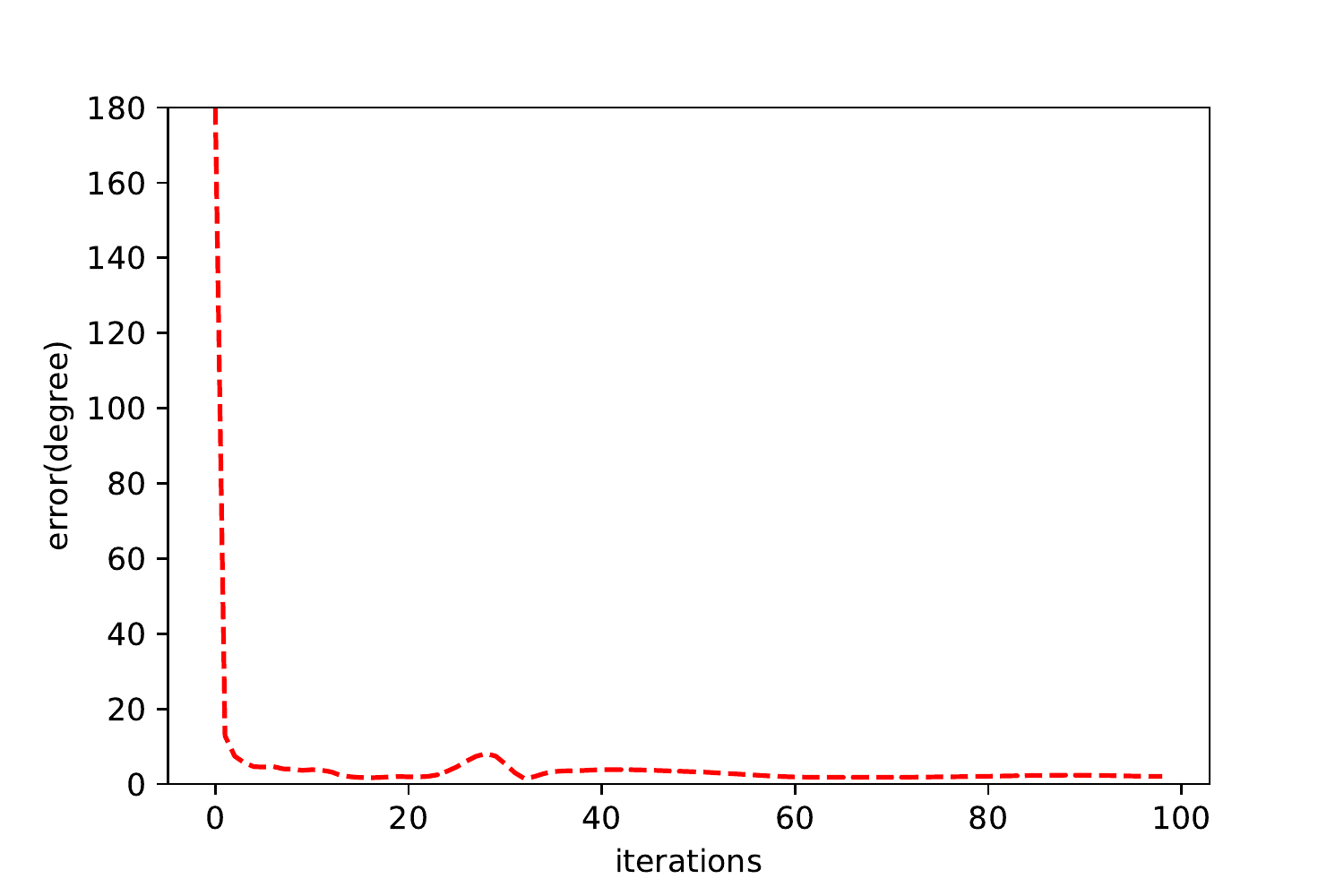}\\
\includegraphics[width=.6\linewidth]{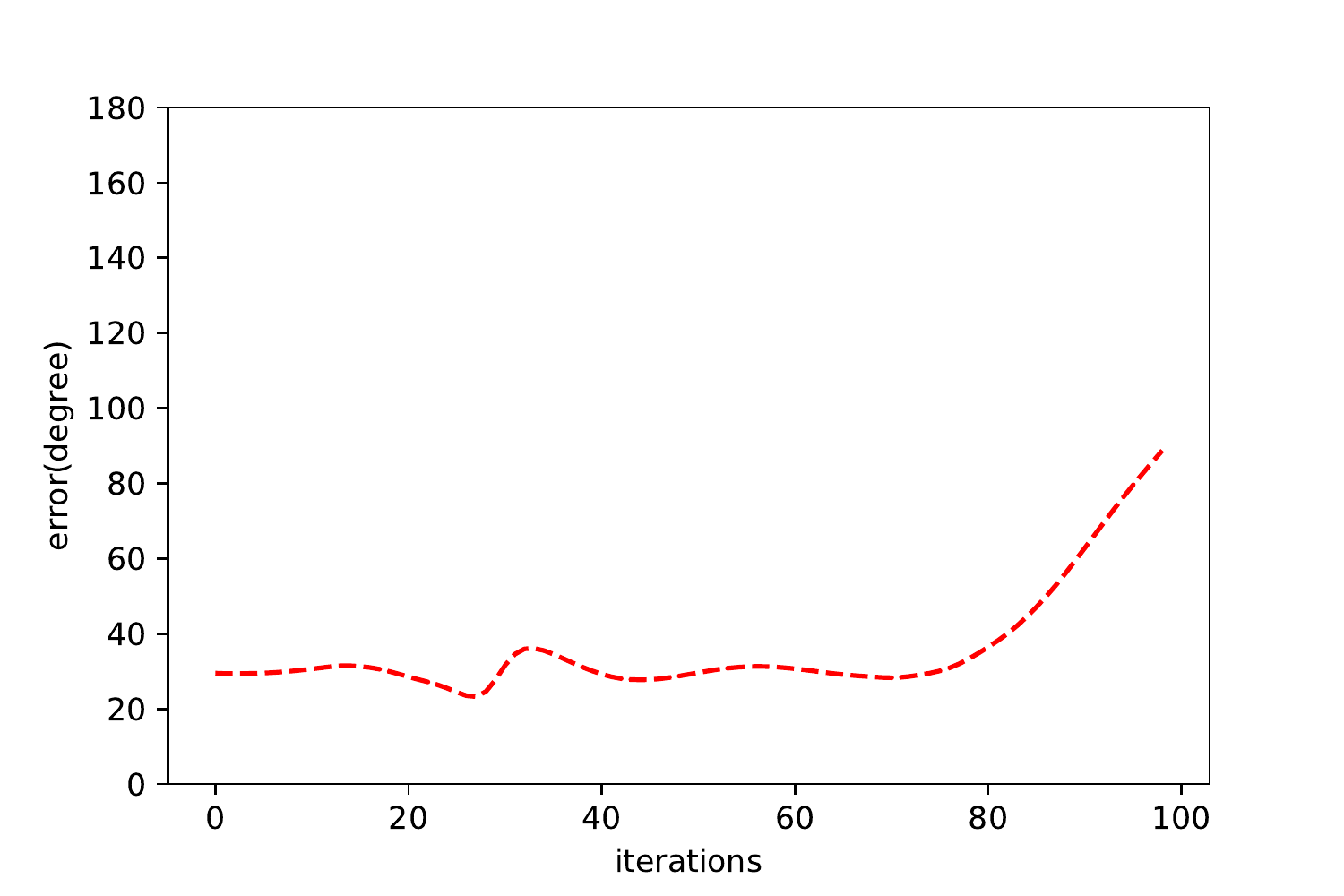}
\caption{Error of angles (in deg) in estimating (top) rotation axes and (bottom) translation vectors over iterations.}

\label{fig:error_KITTI9}
\end{figure}

Next, we evaluated the proposed method with real video frames
from KITTI dataset \cite{Geiger2012CVPR,Geiger2013IJRR}. This is not a non-overlapping calibration problem,
however the ground truth pose is available and we use it for evaluating the accuracy of our method. Figure \ref{fig:error_KITTI9} shows the errors in rotation and translation over iterations.
The error increases as iterations increase, which demonstrates that the proposed method doesn't work well for real video frames of this setting. This is caused by the ego-motion method \cite{Zhou2017} producing pose estimations of different scales in each several frames. We are currently working on tackling this problem.

\section{Conclusions}

We have proposed a method for non-overlapping camera calibration. This is on-line, and use a computation graph to implement the estimation of rotation with incremental SVD, and translation with RLS.
Results with camera poses generated synthetically and obtained from real video frames with PoseNet demonstrate that our method works well in simulation.
Our future work includes experiments and evaluations on real videos taken by non-overlapping cameras.

\section*{Acknowledgements}

This work was supported in part by JSPS KAKENHI grant number JP16H06540.

\bibliographystyle{plain}
\bibliography{myref}

\end{document}